\label{packages}
\documentclass[journal]{IEEEtran}
\usepackage{cite}

\usepackage{graphicx}

\usepackage[cmex10]{amsmath}
\usepackage{amssymb}

\usepackage[ruled,vlined]{algorithm2e}
\SetKwInput{KwIn}{Input}
\SetKwInput{KwOut}{Output}
\SetKwInput{KwIni}{Initial}

\usepackage{array}
\usepackage{multirow}

\usepackage[footnotesize,labelsep=period]{caption}
\usepackage[center,scriptsize,tight]{subfigure}

\usepackage{url}
\usepackage[colorlinks,linkcolor=red,anchorcolor=blue,citecolor=green]{hyperref}

\begin{document}
\title{Long-Range Motion Trajectories Extraction of Articulated Human Using Mesh Evolution}

\author{Yuanyuan~Wu,~Xiaohai~He,~Byeongkeun~Kang,~Haiying~Song~and~Truong~Q.~Nguyen,~\IEEEmembership{Fellow,~IEEE}
      
\thanks{Copyright (c) 2015 IEEE. Personal use of this material is permitted. However, permission to use this material for any other purposes must be obtained from the IEEE by sending a request to pubs-permissions@ieee.org.

This work was supported in part by the National Natural Science Foundation of China (NSFC) under Grant 61471248, NSAF Foundation of China under Grant 11176018 and China Scholarship Council.

Y. Wu and X. He are with the College of Electronics and Information Engineering, Sichuan University, Chengdu 610065, China. (e-mail: yuanyuanwu29@163.com; hxh@scu.edu.cn). H. Song is with the Communication Engineering Department, Chengdu Technological University Chengdu 611730, China. (e-mail: shying08@163.com). B. Kang and T. Q. Nguyen are with Department of Electrical and Computer Engineering, University of California-San Diego, La Jolla, CA 92093, USA (e-mail: bkkang@eng.ucsd.edu; tqn001@eng.ucsd.edu). }}

\maketitle

\label{abstract}
\begin{abstract}
This letter presents a novel approach to extract reliable dense and long-range motion trajectories of articulated human in a video sequence. Compared with existing approaches that emphasize temporal consistency of each tracked point, we also consider the spatial structure of tracked points on the articulated human. We treat points as a set of vertices, and build a triangle mesh to join them in image space. The problem of extracting long-range motion trajectories is changed to the issue of consistency of mesh evolution over time. First, self-occlusion is detected by a novel mesh-based method and an adaptive motion estimation method is proposed to initialize mesh between successive frames. Furthermore, we propose an iterative algorithm to efficiently adjust vertices of mesh for a physically plausible deformation, which can meet the local rigidity of mesh and silhouette constraints. Finally, we compare the proposed method with the state-of-the-art methods on a set of challenging sequences. Evaluations demonstrate that our method achieves favorable performance in terms of both accuracy and integrity of extracted trajectories.
\end{abstract}

\begin{IEEEkeywords}
Motion trajectories, articulated motion, mesh evolution.
\end{IEEEkeywords}
\IEEEpeerreviewmaketitle

\section{Introduction}
 \IEEEPARstart{L}{ong}-range motion trajectories provide more precise and integrated information of a movement and have been extensively used in various applications such as action recognition, motion segmentation, video indexing and retrieval, video manipulation. It is worth to note that only one camera is set in most of the applications, which leads to the loss of much visual information and brings many challenges. Sparse feature trackers such as KLT feature tracker\cite{shi1994good} is often used to extract motion trajectories in video sequence. Moreover, spatially-denser trajectories can be obtained by PV tracker \cite{sand2008particle} and LDOF tracker \cite{sundaram2010dense}. PV tracker builds trajectories by sweeping forward and backward flow fields and also refines motion estimates to enforce long-range consistency. LDOF tracker is based on large displacement optical flow (LDOF) proposed by Brox et al. \cite{brox2009large}. These trackers share one essential criterion that if points are lost possibly due to lighting variation, out of plane rotation, occluded or large displacement, then new points will be added. As a result, points in initial video frame may not be fully tracked throughout the video sequence. However, integrated long-range motion trajectories can be obtained by concatenating frame-to-frame optical flow motion fields, such as Lagrangian particle trajectories (LPT) used in action recognition work \cite{wu2011action}. As discussed in \cite{sand2008particle}, this class of algorithms may cause trajectories drift by error accumulation. In summary, it is challenging to extract both reliable and long-range motion trajectories throughout the whole video sequence.

\begin{figure}[!t]
\centering
\includegraphics[width=0.48\textwidth]{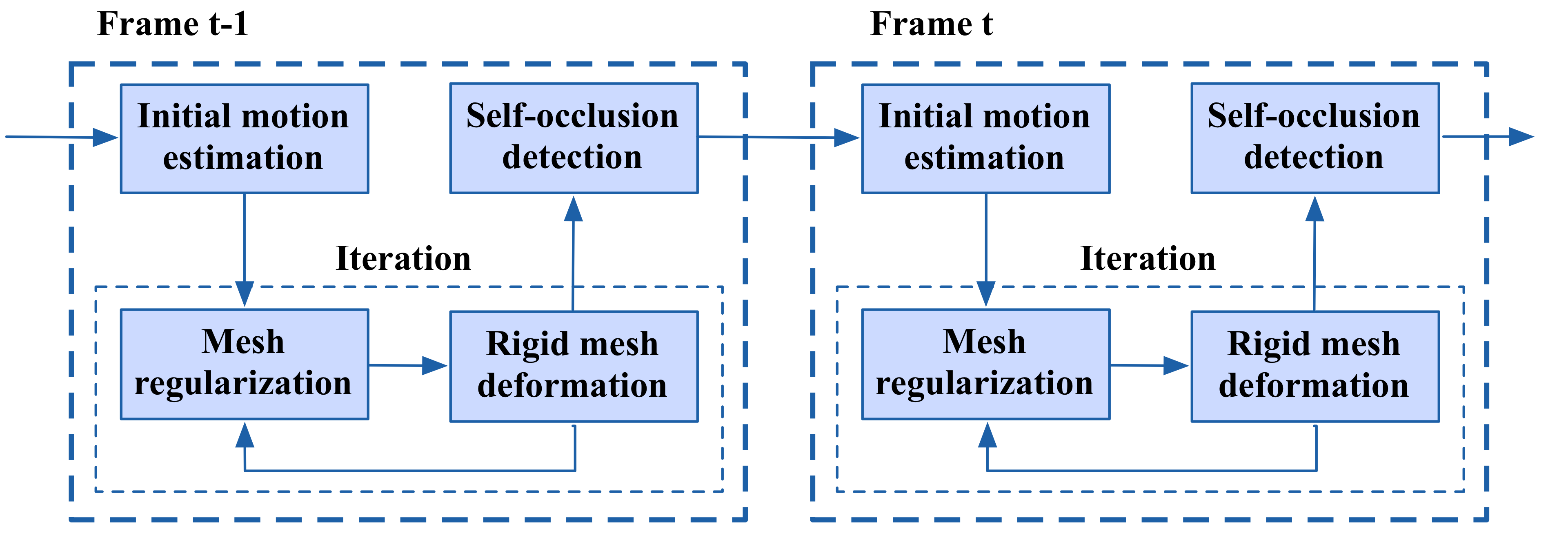}
\caption{Overview of the proposed long-range motion trajectories extraction method from frame $t$-1 to $t$.}
\label{pic: flow chart}
\end{figure}

Our approach is inspired by the work on dense surface tracking in \cite{varanasi2008temporal,cagniart2009iterative}, which both formulate a mesh evolution framework including an iterative mesh deformation step. Differently, \cite{varanasi2008temporal} performs surface-morphing while \cite{cagniart2009iterative} provides local rigidity constraints of a surface in the iterative mesh deformation step. By introducing this mesh evolution framework from 3D space to 2D image plane, we extract long-range motion trajectories effectively. Specifically, self-occlusion is first detected by searching the mesh intersection. Next, vertices in the occlusion region and the non-occlusion region will receive specified motion estimations for propagating to the next frame. Last, vertices are gradually approaching to their actual positions by the iterative mesh deformation step, in which different types of drifted vertices are recognized and regularized, and the local rigidity of the mesh is enforced in an efficient way. In this letter, binary silhouettes of articulated human are utilized to recognize and regularize drifted vertices. Similar to several silhouette-based methods \cite{gorelick2007actions,choudhury2012silhouette,abdelkader2011silhouette,chaaraoui2013silhouette}, the advantages of using silhouettes have been proven in various applications, e.g. human action, gait recognition, etc.. The extraction of silhouettes from a video commonly entails using techniques such as background subtraction. Fig. \ref{pic: flow chart} shows an overview of the proposed long-range motion trajectories extraction method. 
 
Our contribution with respect to methods \cite{varanasi2008temporal,cagniart2009iterative} is that the mesh evolution framework is proposed for monocular-camera set-up. Self-occlusion of object is one inevitable problem in single-view video, so we proposed an effect way to detect the occlusion region. Another problem is that the strategy of mapping after meshing is not applicable in single-view video due to the self-occlusion, so we proposed a strategy of propagating vertices with specified predicted motions. Moreover, some geometric information such as the perspective invariance of surface norms does not extend from surface to silhouette, so we proposed an efficient way to recognize and regularize drifted vertices in 2D. To the best of our knowledge, no previous work has attempted to perform the long-range tracking of articulated human undergoing partial self-occlusion and complicated non-rigid deformations, using silhouettes and mesh evolution in a single-view video.

\section{Proposed trajectories extraction Method}
The input to our system is a monocular video sequence of $M$ frames. The stack of silhouettes $\{S^t\},t\in\{1,\ldots,M\}$ is extracted and $N$ tracked points are sampled uniformly on the reference silhouette $S^1$ by a mesh generator algorithm \cite{persson2004simple,peyre2011numerical}. Let 2-dimensional vector $p_i^t \in \mathbb{R}^2$ denote the position of a tracked point $i$ in frame $t$, then a big matrix $A$ is constructed as follows:
\begin{equation}
A=\begin{array}{@{}r@{}c@{}c@{}c@{}c@{}l@{}}
    &\mathcal{P}^1  & \mathcal{P}^2  & \mathcal{P}^t & \mathcal{P}^M   \\
    \left.\begin{array}
    {c} \mathcal{T}_1 \\\mathcal{T}_2 \\\mathcal{T}_i  \\\mathcal{T}_N  \end{array}\right(
                    & \begin{array}{c} p_1^1 \\ p_2^1\\ \vdots\\ p_N^1 \end{array}
                    & \begin{array}{c} p_1^2 \\ p_2^2 \\ \vdots \\ p_N^2 \end{array}
                    & \begin{array}{c} \ldots \\ \ldots \\ p_i^t \\ \ldots \end{array}
                    & \begin{array}{c}  p_1^M  \\ p_2^M \\ \vdots \\ p_N^M \end{array}
                    & \left)\begin{array}{c} \\ \\ \\  \\ \end{array}\right.
  \end{array}
\end{equation}
 
Note that each row of matrix $A$ is a representation of one fully tracked trajectory $\mathcal{T}_i,i\in\{1,\ldots,N\}$. The objective of our approach is to extract a reliable set of long-range trajectories $\{\mathcal{T}_i\}$. From an alternate point-of-view, $N$ track points are physically belonging to a human undergoing articulated motion. Therefore, each column of matrix $A$ is one instant pose of articulated human which is assumed to share the same topology. We consider a planar triangle mesh $\mathcal{G}^t(\mathcal{V},\mathcal{E},\mathcal{F},\mathcal{P}^t)$ which represents a column of matrix $A$, where $\mathcal{V} = \{ 1, \ldots ,N \}$ is the set of vertices, $\mathcal{E} = \{(i,j),i,j\in\mathcal{V}\}$ is the set of edges,  $\mathcal{F} = \{(i,j,k),i,j,k\in\mathcal{V}\}$ is the set of faces, $\mathcal{P}^t = \{ p_1^t, \ldots ,p_N^t \}$ is the set of vertices positions. We assume that all meshes $\{\mathcal{G}^t\}$ share the same topology $(\mathcal{V},\mathcal{E},\mathcal{F})$ but vary at vertex positions $\mathcal{P}^t $. Therefore, the trajectories extraction problem is casted as mesh evolution over time. i.e.
\begin{equation}
\label{eq:deformation}
\mathcal{G}^1( {\mathcal{V},\mathcal{E},\mathcal{F},\mathcal{P}^1} ) \to \mathcal{G}^t( {\mathcal{V},\mathcal{E},\mathcal{F},\mathcal{P}^t})
\end{equation}

\subsection{Self-Occlusion Detection}
\begin{figure}[!h]
\centering
\subfigure[]{
\includegraphics[width=0.095\textwidth]{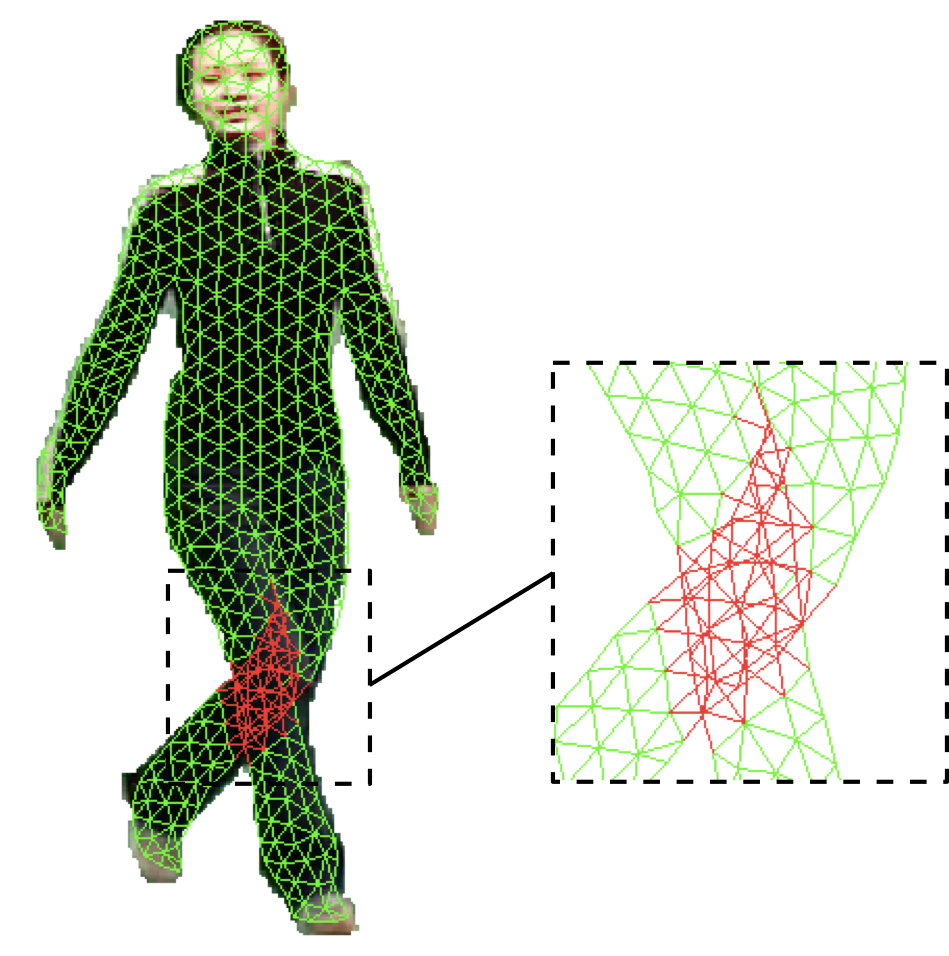}
\label{subpic: crossing}
}
\subfigure[]{
\includegraphics[width=0.1\textwidth]{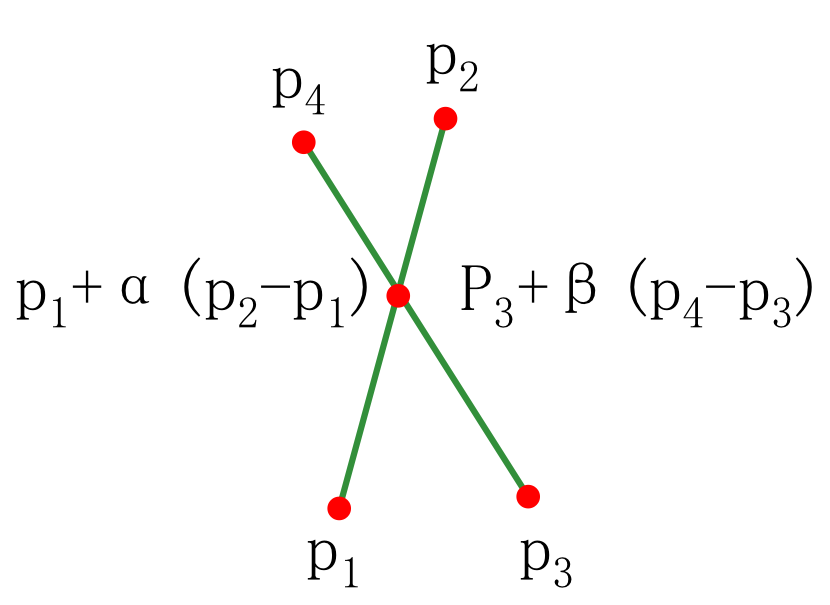}
\label{subpic: theory}
}
\caption{An example of detecting self-occlusion in one frame of $Walking$ sequence. (a) intersected edges in occlusion region are colored in red, (b) illustration of two intersected edges in the mesh.}
\label{pic: occlusion-detection}
\end{figure}

Self-occlusion is commonly occurring between moving torso and swinging limbs undergoing articulated motions. By taking the advantage of the deformed mesh, we detect the occlusion region by finding intersected edges of the mesh. As illustrated in Fig. \ref{subpic: crossing}, during the leg crossing motion, two components of mesh intersect in the occlusion region which is highlighted in red color. In computational geometry, this is a line segment intersection problem which supplies a list of line segments in the Euclidean plane and asks whether any two of them intersect. As illustrated in Fig. \ref{subpic: theory}, suppose the two line segments run from $p_1$ to $p_2$ and from $p_3$ to $p_4$. Then any point on the first line is represented as $p_1+\alpha(p_2 - p_1)$ and similarly $p_3+\beta(p_4 - p_3)$ is for any point on the second line, where $\alpha$ and $\beta$ are scalar parameters. The two line segments intersect if we can find $\alpha$ and $\beta$ such that:
\begin{equation}
p_1+\alpha(p_2 - p_1)=p_3+\beta(p_4 - p_3)
\end{equation}
Cross both sides with $p_4-p_3$ and $p_2-p_1$ separately, solving for $\alpha$ and $\beta$:
\begin{equation}
\alpha=\|(p_3-p_1)\times(p_4-p_3)\|/\|(p_2-p_1)\times(p_4-p_3)\|
\end{equation}
\begin{equation}
\beta=\|(p_1-p_3)\times(p_2-p_1)\|/\|(p_4-p_3)\times(p_2-p_1)\|
\end{equation}

If the denominator $\|(p_2-p_1)\times(p_4-p_3) \|= 0$, then the two lines are parallel or collinear. Otherwise, if $\|(p_2-p_1)\times(p_4-p_3) \|\ne 0$ as well as $0<\alpha<1$ and $0<\beta<1$, then two lines intersect. Therefore, intersected edges are found in the mesh and corresponding vertices are identified in occlusion region.

\subsection{Initial Motion Estimation}
In order to propagate mesh $\mathcal{G}^{t-1}$ to $\mathcal{G}^{t}$ in the next frame for a reliable initial guess, we propose to estimate the vertices of $\mathcal{G}^{t}$ through large displacement optical flow (LDOF) \cite{brox2009large}, polynomial curve fitting, and patch-based average filtering. LDOF as a recent successful optical flow method, particularly approach the problematic of estimation of articulated human motion. However, it does not solve occlusion problem like other optical flow methods. Therefore, an adaptive method is proposed to estimate motion vectors of vertices of $\mathcal{G}^{t-1}$ in different image regions: For a vertex $p_i^{t-1}$ in non-occlusion region, we perform bicubic spline interpolation of LDOF motion vectors to get the motion vector $u_i^{t-1}$. For a vertex $p_i^{t-1}$ in occlusion region, we perform a second-order polynomial curve fitting to construct vertex $p_i^{t}$ within the range of a discrete set of previous five positions. Specifically, the fitting model is $Y_i=BX_t$, where $B = \left[ {\begin{array}{*{20}{c}}
{{a_1}}&{{b_1}}&{{c_1}}\\
{{a_2}}&{{b_2}}&{{c_2}}
\end{array}} \right]$ is the unknown coefficients matrix, $X_t$ and $Y_i$ respectively are input and output matrices, i.e. $X_t = [x_{t-1}\ x_{t-2}\ x_{t-3}\ x_{t-4}\ x_{t-5}\ ], \ x_t = [t^2\ t\ 1]^T$, $Y_i = [p_i^{t-1}\ p_i^{t-2}\ p_i^{t-3}\ p_i^{t-4}\ p_i^{t-5}]$. Therefore, the solution of coefficients matrix is $B=Y_iX_t^{T}(X_t{X_t}^T)^{-1}$ and the estimated motion vector is 
\begin{equation}
u_i^{t-1}= Bx_t - p_i^{t-1}
\end{equation}

Moreover, in order to handle the observation noise, we apply a patch-based average filter to obtain smoothing result of motion vectors. Here, a patch is denoted as the set of vertex $i$ and its adjacent vertices, i.e. $N(i) = \{i\} \cup \{j:(i,j) \in \mathcal{E}\}$. $|N(i)|$ defines the number of vertices in patch $N(i)$. Specifically, the proposed motion estimation method is defined as 
\begin{equation}
 p_{i;Initial}^t = p_i^{t-1} + \frac{1}{|N(i)|} \sum\limits_{j\in N(i)} u_j^{t-1}
\end{equation}

\subsection{Iterative Mesh Deformation}
The previous step provides a reasonable initialization of vertex positions at frame $t$ by taking into account the self-occlusion problem. Further refinement is necessary to solve the drift problem which can be caused by non-rigid motion, large displacement, variations in appearance and light, and interference from ambiguous textures. An iterative solution of mesh regularization and rigid mesh deformation is proposed to get the optimal estimation result $\hat p_i^t(k)$ with the initialization of $\hat p_i^t (0) = p_{i;Initial}^t$ , where $k$ is the iteration number. We then define the energy function as follows:
\begin{equation}
f(k) = \sum\limits_{i = 1}^N {\left\| {\hat p_i^t(k) - \hat p_i^t(k-1)} \right\|^2} 
\end{equation}
In order to reduce the effect of noise and various value range of data, the energy function is first normalized by linear normalization, then it is fitted by the power function ($y = a{x^b}$). We then define the iteration stopping criteria by the fitted energy function as follows ($\theta $ is set as 0.003 in our experiments):

\begin{equation}
\left| {\hat f(k) - \hat f(k-1)} \right| < \theta
\label{eq: stop}
\end{equation}

\begin{figure}[!t]
\centering
\subfigure[]{
\includegraphics[width=0.08\textwidth]{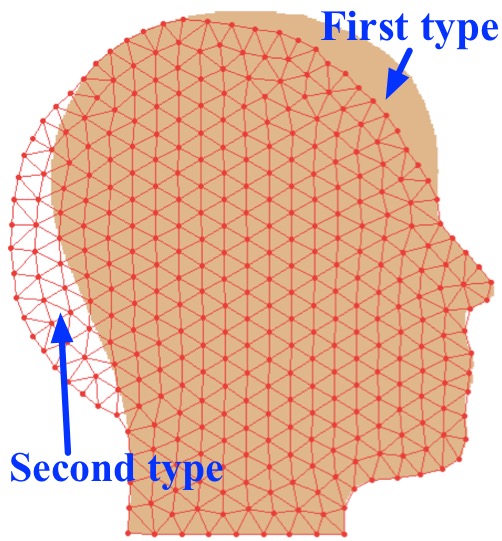}
\label{subpic: original_mesh}
}
\subfigure[]{
\includegraphics[width=0.08\textwidth]{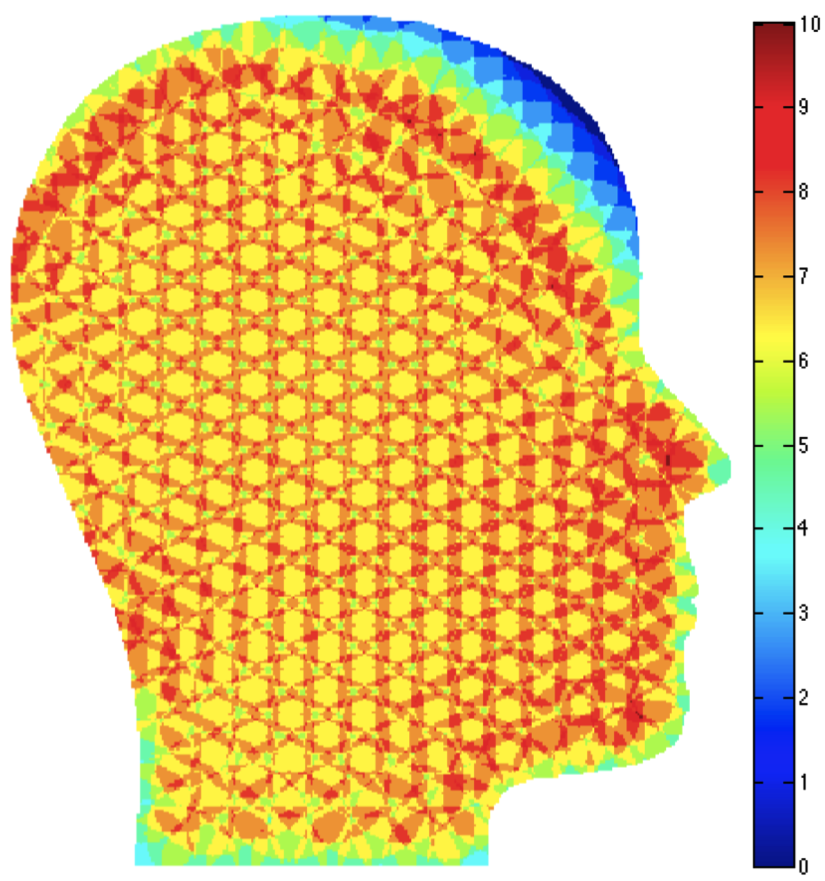}
\label{subpic: density}
}
\subfigure[]{
\includegraphics[width=0.08\textwidth]{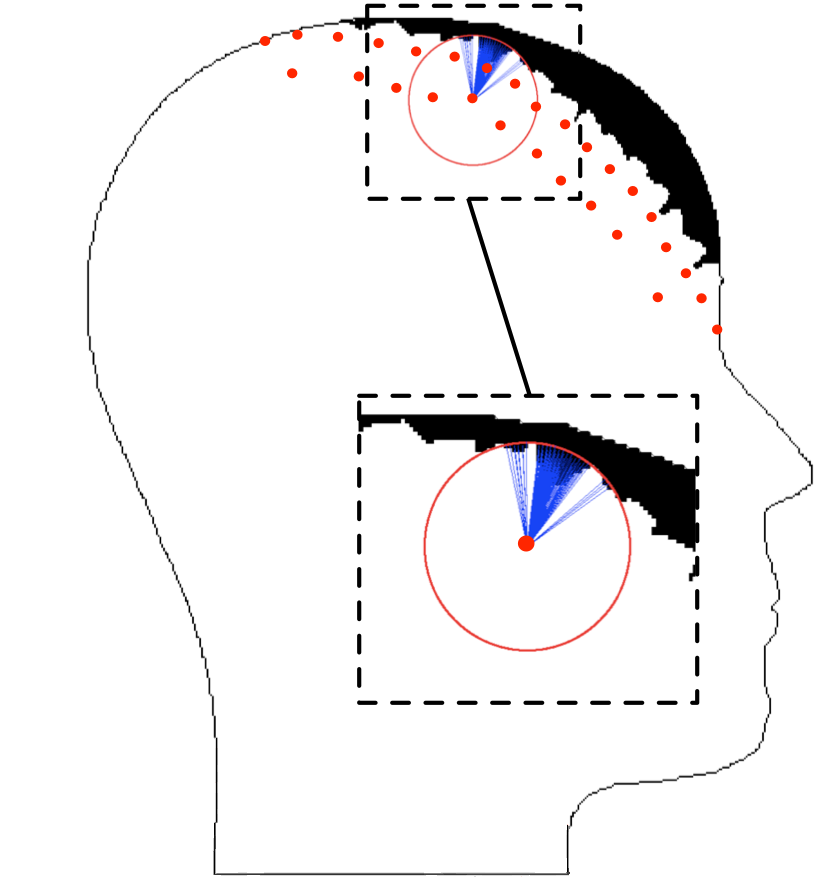}
\label{subpic: empty}
}
\subfigure[]{
\includegraphics[width=0.08\textwidth]{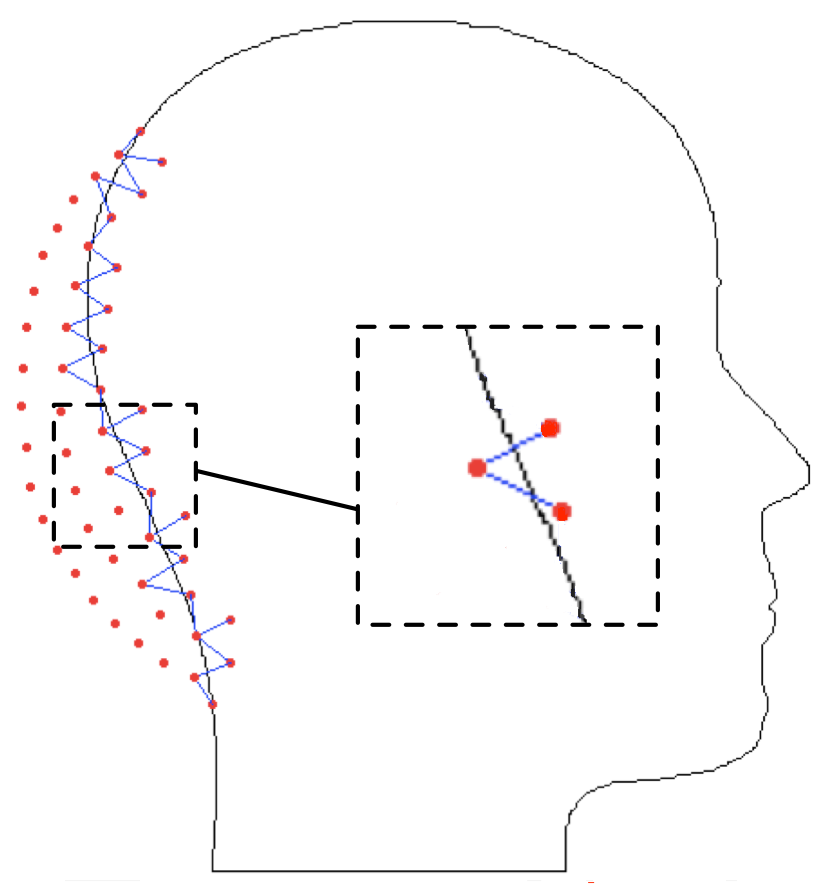}
\label{subpic: out}
}\vspace{-10pt}
\subfigure[]{
\includegraphics[width=0.08\textwidth]{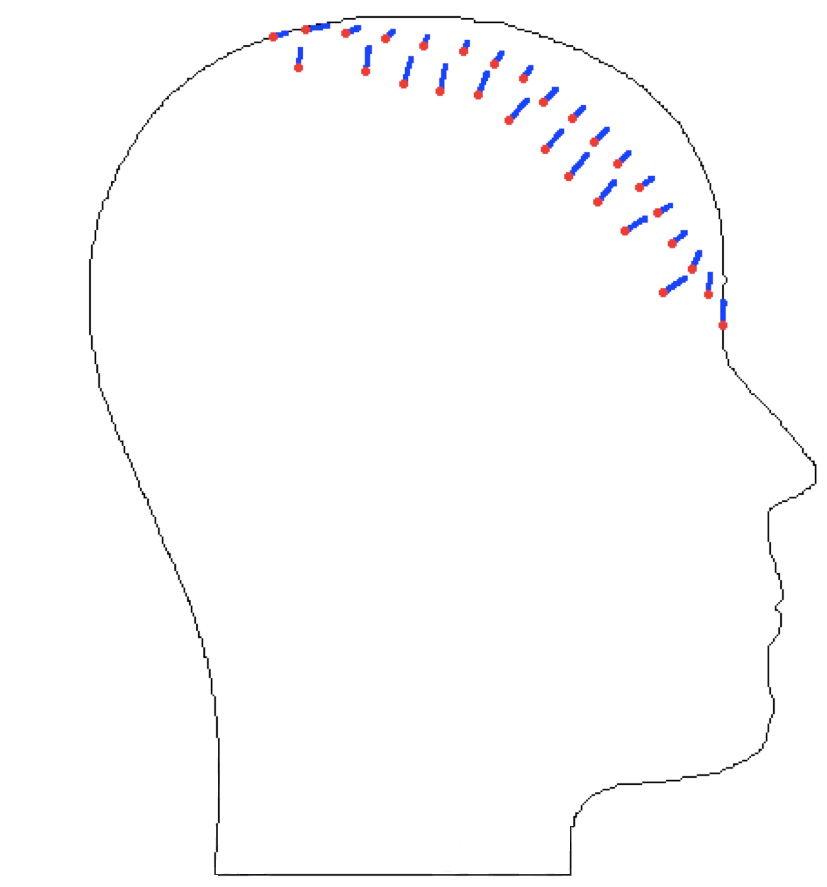}
\label{subpic: out1}
}
\subfigure[]{
\includegraphics[width=0.08\textwidth]{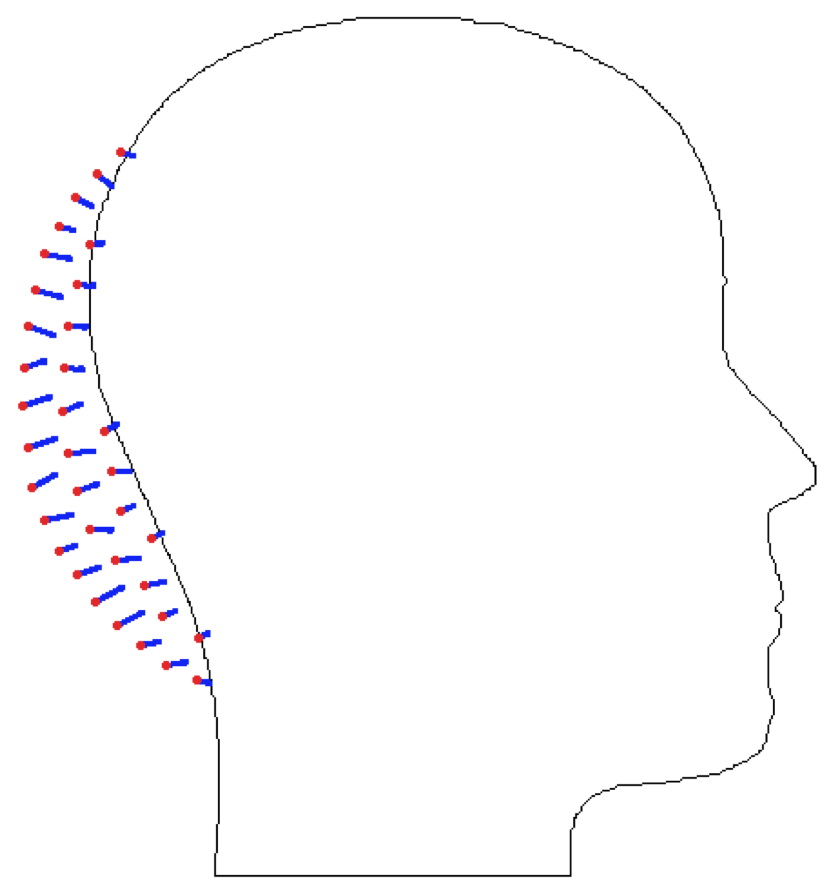}
\label{subpic: out2}
}
\caption{Illustration of mesh regularization process. (a) initial mesh and the silhouette, (b) vertex density map, (c) the regularization of first type of drifted vertices, (d) the regularization of second type of drifted vertices, (e) and (f) are displacement vectors of regularized vertices.}
\label{pic: mesh regular}
\end{figure}

\subsubsection{Mesh Regularization}

When vertices drift away from their actual positions, the constructed mesh no longer meets the silhouette constraint. Typically, there are two types of drift vertices: the first type is when vertices do not reach the actual positions, which leads to the blank of silhouette; the second type is when vertices are beyond the range of the silhouette, as shown in Fig. \ref{subpic: original_mesh}. To predict the target position, drifted vertices are gradually regularized toward  the blank of silhouette and away from non-silhouette area. First, we compute the vertex density map, which is a measurement of vertices per unit area (within the radius of the longest edge of reference mesh $\mathcal{G}^1$), as shown in Fig. \ref{subpic: density}. By giving a threshold, the blank of silhouette is simply labeled and expressed as a set of pixel points $Q=\{q_1,q_2,...\}$, as shown as black region in Fig. \ref{subpic: empty}. If a subset $Q_i \subset Q$ is within the unit area of a vertex $i$, we denote the vertex $i$ as the first type of drifted vertices ($\mathcal V_1$), and  will predict its target position from the pixel points in $Q_i$. As shown in Fig. \ref{subpic: out}, if a vertex is beyond the range of silhouette, we denote it as the second type of drifted vertices ($\mathcal V_2$) and predict its target position from support adjacent vertices which are denoted as $N_i = N(i) \cap (\mathcal V{\rm{\backslash }}{{\rm{\mathcal V}}_2})$. Note that a potential issue could occur where a patch of vertices are all second type of drifted vertices, that is, $N(i) \subset \mathcal V_2$ and the set $N_i = null$. Therefore, we predict the target positions for the second type of drifted vertices in a batch process. The predicted batch of vertices will be removed from set $\mathcal V_2$, and keep predicting left vertices until $\mathcal V_2$ is empty. We can finally regularize the target position as follows: 
\[p_{i;Reg}^t (k) =\]
\begin{eqnarray}
\begin{split}
\begin{cases}
 \lambda \hat p_{i}^t(k-1) + (1-\lambda) \frac{1}{|Q_i|} {\sum\limits_ {q_j\in Q_i} q_j}    & if \ i\in \mathcal V_1  \\
 \lambda \hat p_{i}^t(k-1)  + (1-\lambda) \frac{1}{| {N_i} |} {\sum\limits_ {j\in N_i}\hat p_{j}^t(k-1) } & if \ i\in \mathcal V_2\\
\hat p_{i}^t(k-1)  & else \\
\end{cases}
\end{split}
\end{eqnarray}

Here, $|Q_i|$ and $|N_i|$ are the number of elements of set $Q_i$ and $N_i$ respectively. The $\lambda$ term balances the influence of original point and points in support domain; controls the regularization pace. In practice $\lambda = 2/3$ was used for all experiments. Fig. \ref{subpic: out1} and \ref{subpic: out2} show the results of mesh regularization.

\subsubsection{Local Rigid Deformation}

To preserve the local rigidity of the deformed mesh, we map the patches to a global coordinate system via per-patch rigid transformations, here the rigid transformation is equivalent to an affine transformation in 2D image plane. As described in simulation (\ref{eq:deformation}), we would like to compute the rigid transformation of a reference patch in $\mathcal{P}^1$ to best conform it to the corresponding patch in $\mathcal{P}^t$, such that:
\begin{equation}
(R_i,T_i )\leftarrow \arg \min \sum\limits_{j\in N(i)} \|p_{j;Reg}^t(k) -(R_i p_j^1+T_i)\|^2
\end{equation}
where $R_i$ is the $2\times2$ rigid transformation matrix and $T_i$ is the translation vector. This is an instance of procrustes problem, which can be solved by procrustes analysis\cite{gower2004procrustes}. Instead of simply using the rigid transformation of patch $N(i)$, we also consider the rigid transformations from the neighboring patches $\{N(j)\},j\in N(i)$. This procedure preserves the local rigidity of mesh deformation better. The vertex position is defined as
\begin{equation}
\label{eq:RD}
p_{i;RD}^t(k) = \frac{1}{| {N(i)} |}\sum\limits_{j \in N(i)} {({R_jp_i^1 + T_j})} 
\end{equation}

After the mesh regularization and local rigid deformation of mesh, one iteration ends and the next iteration begins with the updated position, i.e. $\hat p_i^{t}(k)  = p_{i;RD}^t(k)$. The iteration stops when satisfy the stopping criteria in equation \ref{eq: stop}.

\section{Experiments}
\begin{figure*}[!thb]
\centering
\hspace{-.15in}
\subfigure[Walk]{
\includegraphics[width=0.19\textwidth]
{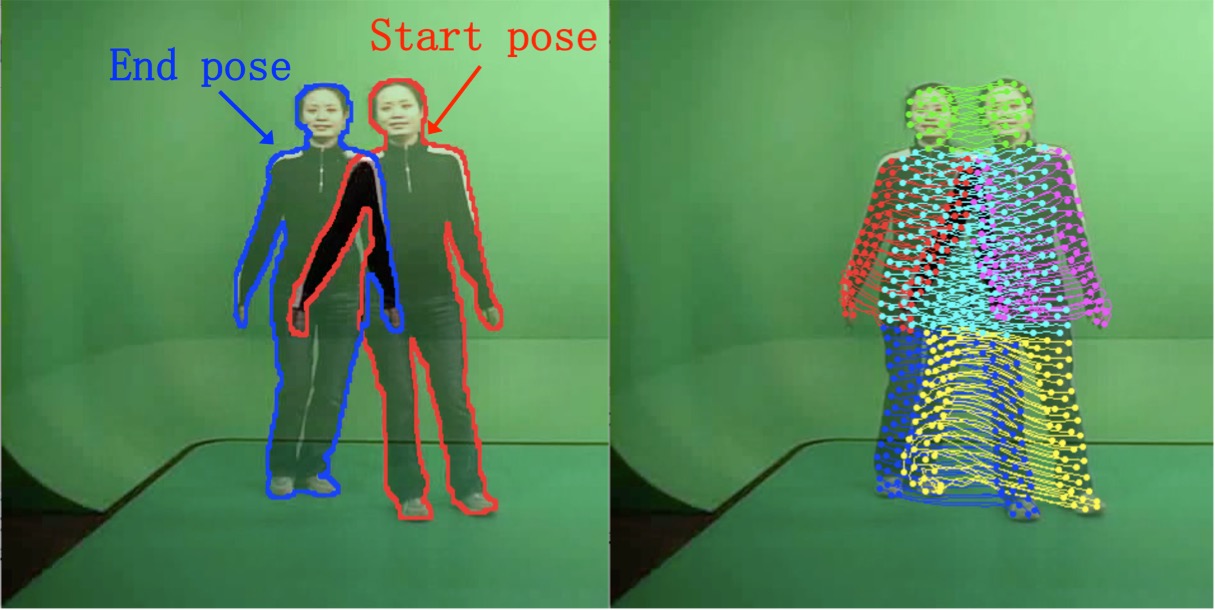}
\label{subpic:walking}
}
\hspace{-.15in}
\subfigure[Wheel]{
\includegraphics[width=0.19\textwidth]
{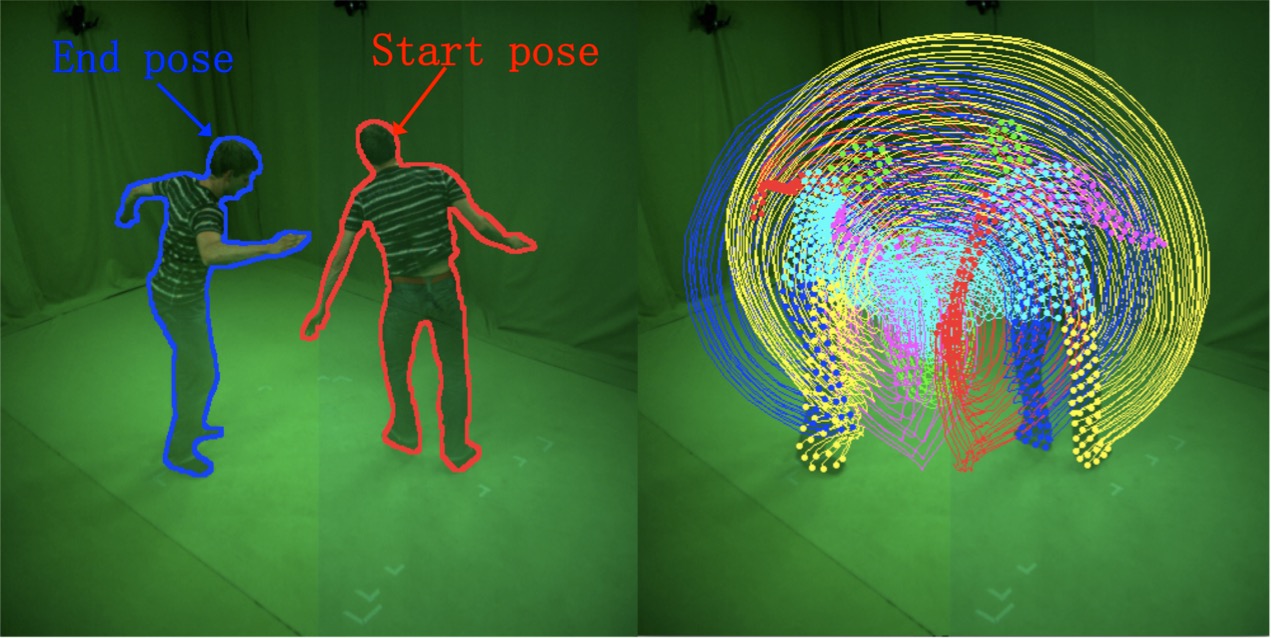}
\label{subpic: wheeling}
}
\hspace{-.15in}
\subfigure[Handstand]{
\includegraphics[width=0.19\textwidth]
{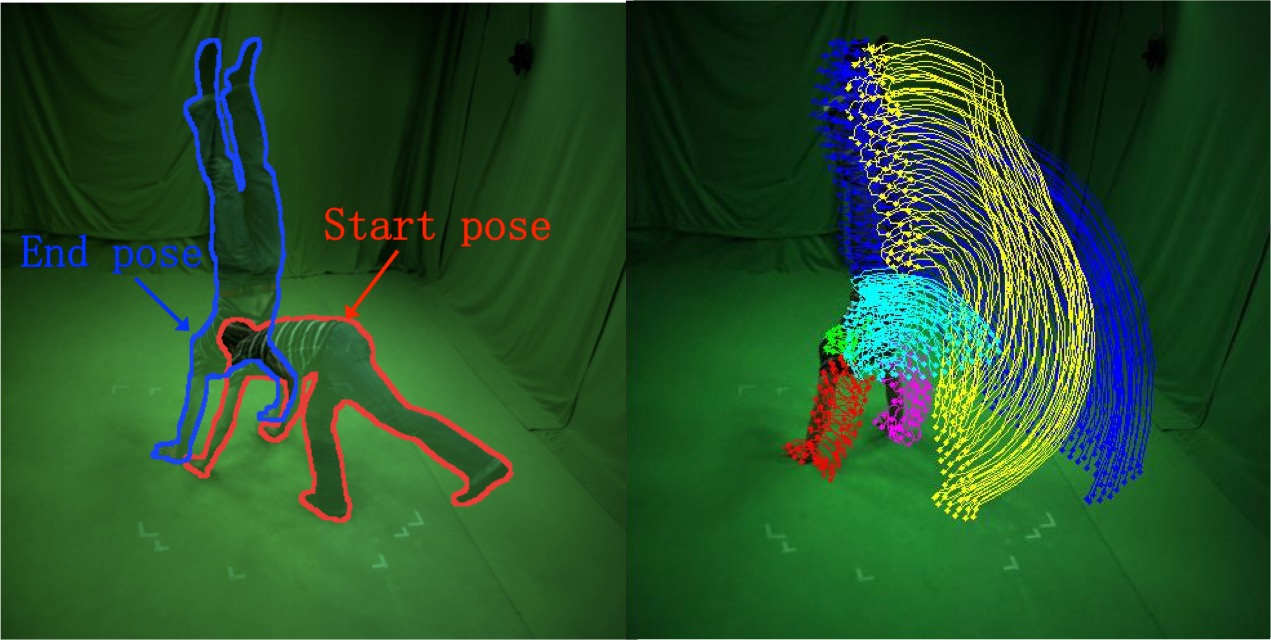}
\label{subpic: handstanding}
}
\hspace{-.15in}
\subfigure[Dance]{
\includegraphics[width=0.19\textwidth]
{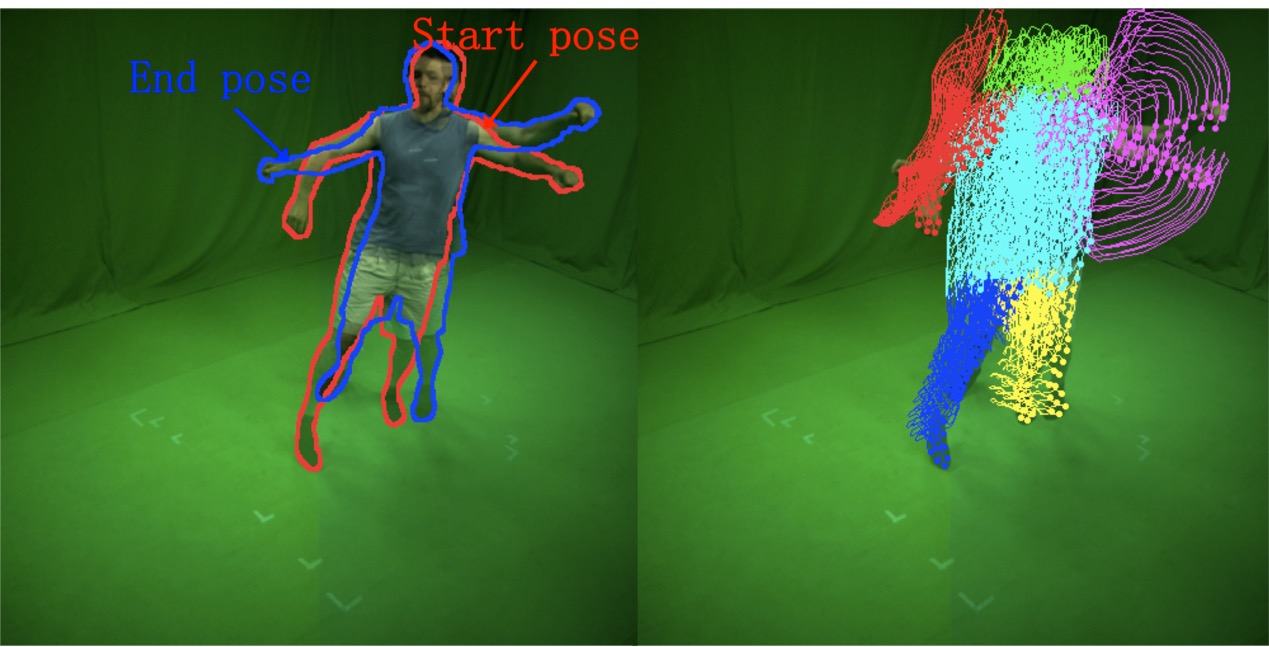}
\label{subpic: dancing}
}
\hspace{-.15in}
\subfigure[Skirt]{
\includegraphics[width=0.19\textwidth]{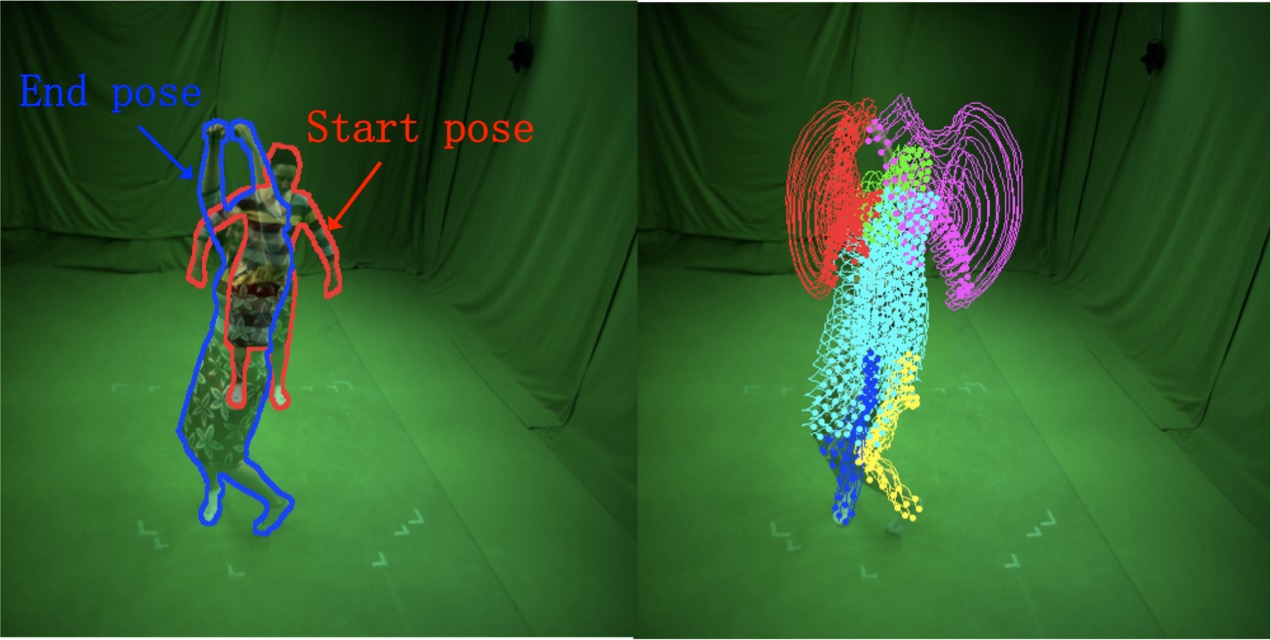}
\label{subpic: skirt}
}
\hspace{-.15in}
\caption{Results of the proposed method on five sequences. The body parts are best viewed in color.}
\label{pic: long trajectories}
\end{figure*}

\subsection{Datasets and Baselines}
To evaluate the efficiency of the proposed method, five challenging sequences from \cite{gall2009motion,wang2006modular}  and Weizmann Human Action Dataset \cite{ActionsAsSpaceTimeShapes_iccv05} are used. The challenges of these videos include pose change, self-occlusion, rapid movement, and scale variation. Our method is also compared with some state-of-the-art motion trajectories extraction algorithms including KLT tracker\cite{shi1994good}, PV tracker\cite{sand2008particle}, LDOF tracker \cite{sundaram2010dense} and LPT \cite{wu2011action}. Their source codes are provided by the authors and the parameters are tuned to achieve the best results.

\subsection{Long-Range Motion Trajectories Extraction}
\begin{figure}[!hb]
\centering
\hspace{-.15in}
\subfigure[Wave1]{
\includegraphics[width=0.11\textwidth]
{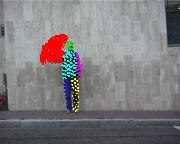}
}
\hspace{-.15in}
\subfigure[Jack]{
\includegraphics[width=0.11\textwidth]
{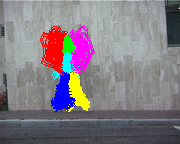}
}
\hspace{-.15in}
\subfigure[Run]{
\includegraphics[width=0.11\textwidth]
{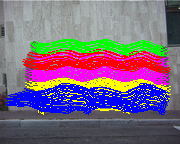}
}
\hspace{-.15in}
\subfigure[Jump]{
\includegraphics[width=0.11\textwidth]
{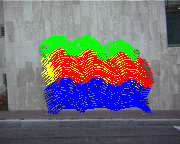}
}

\caption{Results of the proposed method on Weizmann Human Action Dataset. The body parts are best viewed in color.}
\label{pic: weizmann}
\end{figure}

Fig. \ref{pic: long trajectories} illustrates the epitome of five sequences and the extracted long-range motion trajectories (the longest motion trajectory in time is 141 frames from the Skirt sequence). Each sequence has its own characteristics. In the sequence $Walking$, lightly foreshortening and self-occlusion have occurred when the woman moved her left leg diagonal backward followed by her right leg moving. The sequences $Wheeling$ and $Hand standing$ recorded a complete wheeling action and hand standing action respectively, fast movement and out-of-plane rotation are the main challenges. The sequence $Dancing$ contained complex pose change, foreshortening and self-occlusion. In sequence $Skirt$, the women moved forward with her arms lift and then turned sideways, undergoing scale variation and out-of-plane rotation. The proposed method achieved robust performance over these challenging sequences. We also test our approach on the Weizmann Human Action Dataset \cite{ActionsAsSpaceTimeShapes_iccv05}, and some of the results are shown in Fig. \ref{pic: weizmann}. The visual results can be found in our project website \url{http://videoprocessing.ucsd.edu/~yuanyuan/trajectores.html}.

\subsection{Performance Comparison}
\begin{figure}[!h]
\centering
\subfigure[Walk]{
\includegraphics[width=0.48\textwidth]{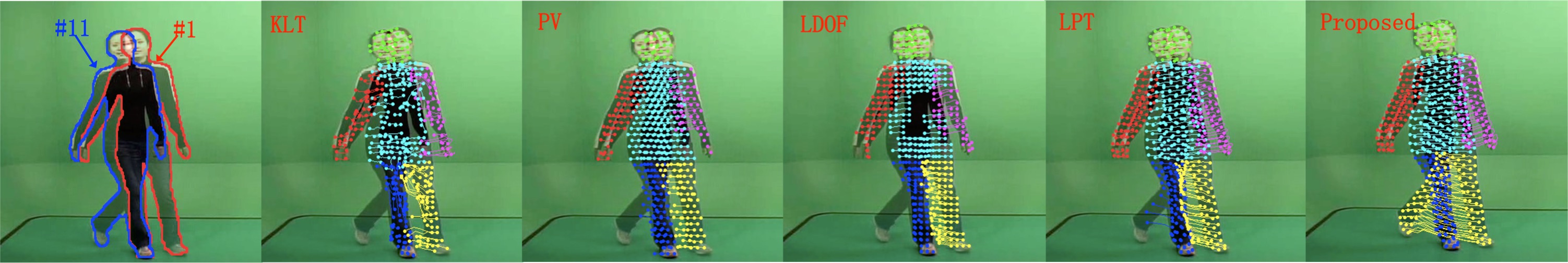} 
\label{subpic: occ}%
}\vspace{-10pt}
\subfigure[Wheel]{
\includegraphics[width=0.48\textwidth]{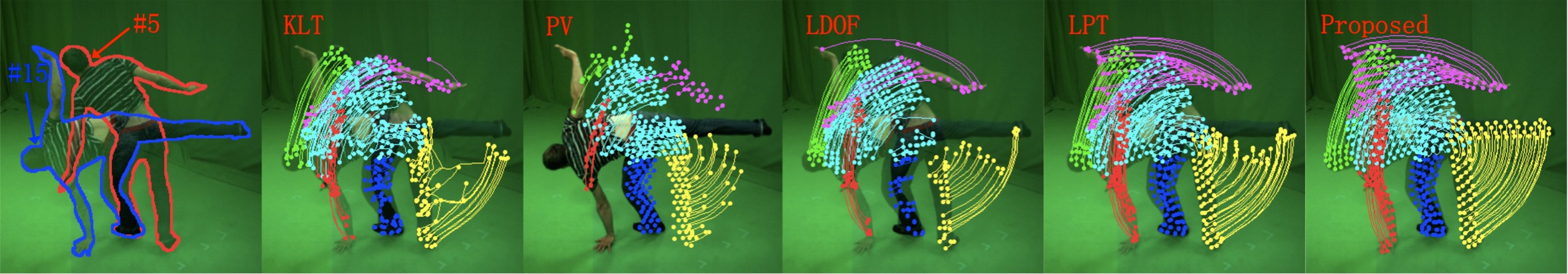}
\label{subpic: fast}%
}
\caption{Sub-trajectories of KLT, PV, LDOF, LPT and the proposed method on two challenging sequences.}
\label{pic: sub trajectories}
\end{figure}

To evaluate the accuracy of extracted motion trajectories by the state-of-the-art methods and the proposed method, we illustrate the visual comparisons in Fig. \ref{pic: sub trajectories}, where self-occlusion and fast movement happens in sequence $Walking$ and sequence $Wheeling$. It is observed that an abundance of points on the leg drifted away or stopped tracking due to self-occlusion and fast movement when using other four methods while the proposed method tracked dense points accurately. 

In this paper, the percentage of tracking length in time is  computed to evaluate the integrity of extracted motion trajectories. From Table \ref{table} we can observe that the average percentage of tracking length in time by KLT, PV, LDOF algorithms are less than 100$\%$, that means these algorithms can not continually track dense points throughout all the five sequences. In contrast, integrated trajectories are obtained by LPT and the proposed method. 

\begin{table}[!h]
\centering  
\caption{The average percentage of tracking length in time.}
\label{table}
\begin{tabular}{m{30pt}m{25pt}m{25pt}m{25pt}m{25pt}m{35pt}}
\hline
Video &KLT(\%) &PV(\%)  &LDOF(\%) &LPT(\%) &Proposed(\%) \\ \hline 
Walk &57.4 &67.4 &61.6 &100 &100\\  
Wheel &35.1 &18.9 &23.1 &100 &100 \\
Handstand &42.1 &34.8 &21.4 &100 &100 \\
Dance &83.0 &43.8 &34.0  &100 &100\\
Skirt &99 &79.1 &27.5 &100 &100 \\ \hline
\end{tabular}
\end{table}

\begin{figure}[!hb]
\centering
\hspace{-10pt}
\subfigure[Walk]{
\includegraphics[width=0.095\textwidth]
{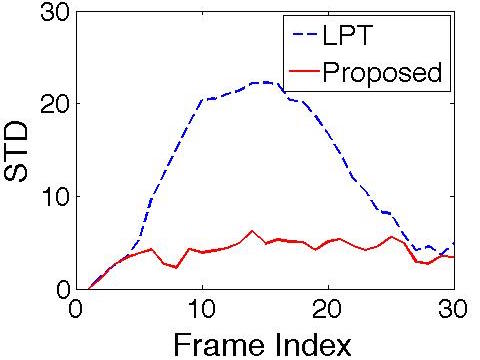}
}\hspace{-10pt}
\subfigure[Wheel]{
\includegraphics[width=0.095\textwidth]
{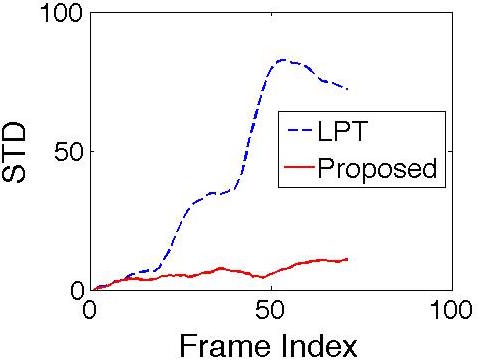}
}\hspace{-10pt}
\subfigure[Handstand]{
\includegraphics[width=0.095\textwidth]
{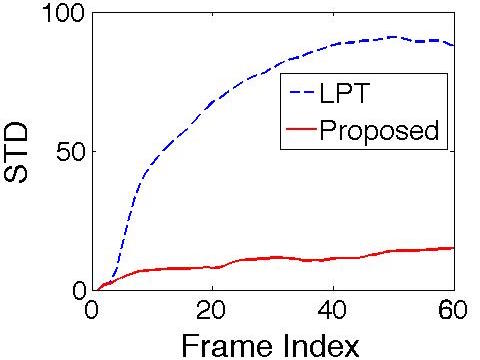}
}\hspace{-10pt}
\subfigure[Dance]{
\includegraphics[width=0.095\textwidth]
{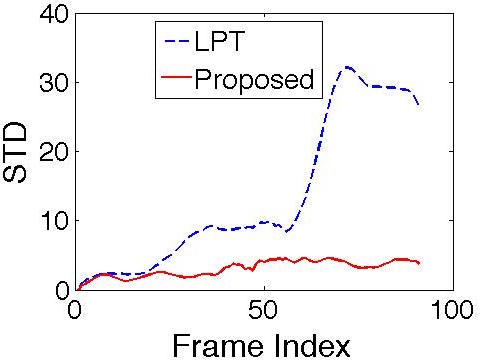}
}\hspace{-10pt}
\subfigure[Skirt]{
\includegraphics[width=0.095\textwidth]
{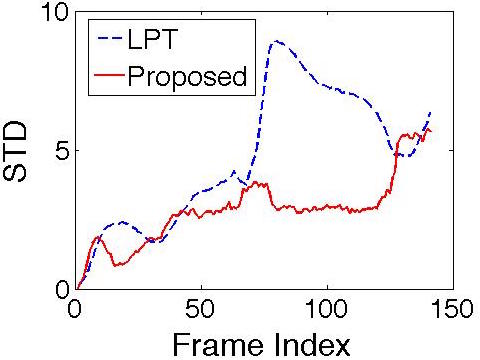}
}
\caption{The standard deviation of offset distances from extracted joint center positions to benchmarks in every frame of five sequences.}
\label{pic: error }
\end{figure}

To further evaluate the accuracy of integrated motion trajectories extracted by LPT and the proposed method, we compute the tracking error based on the provided benchmarks of joint center positions in every frame \cite{gall2009motion,wang2006modular}. Fig. \ref{pic: error } presents the standard deviation of the offset distances in every frame of five sequences. It is observed that the proposed method outperforms LPT with smaller value of the standard deviation of offset distances. It is worth to point out that taking advantages of silhouettes may be the main reason that makes the proposed method superior to LPT. Silhouette constraints play an important role in recognizing and regularizing drifted vertices, therefore avoiding the accumulation of errors during the tracking.

\section{Conclusion}
This letter presents a novel effective and reliable long-range motion trajectories extraction method based on mesh evolution and silhouette constraints. Experiments on challenging video sequences show that the proposed method guarantees the integrity and accuracy of dense points tracking and performs better than several state-of-the-art methods. Since the proposed method is applicable to partial occlusion not full occlusion, it is limited to some challenge actions like spinning around and severe shape deformation. The proposed method is suitable for applications where accuracy of the motion estimation is vital.  

\footnotesize
\bibliographystyle{IEEEbib}
\bibliography{reference}

\end{document}